# AUDIO-BASED TACTILE HUMAN-ROBOT INTERACTION RECOGNITION

A. Yepes[1,2] and M. Charbonneau[1]
[1] Department of Mechanical and Manufacturing Engineering, University of Calgary, Calgary, Canada.
[2] Department of Bioengineering, Universidad de Antioquia, Medellín, Colombia
Email: antonia.yepes@udea.edu.co, marie.charbonneau@ucalgary.ca

## INTRODUCTION

This study explores the use of microphones placed on a robot's body to detect tactile interactions via sounds produced when the hard shell of the robot is touched. Traditional methods using joint torque sensors or 6-axis force/torque sensors are effective but costly and generally only provide a broad indication of contact location [1]. This limitation restricts precise evaluation of the contact's nature (e.g., tapping, knocking, rubbing, stroking, scratching, pressing), which is critical for developing appropriate robot responses. Related work investigated touch classification on a subset of the types of contact mentioned above [2], and touch localization on a rigid robot part [3] based on piezoelectric contact microphone signals. Here, we investigate the viability of using traditional microphones based on air vibration.

## METHODS

Two Adafruit I2S MEMS microphones integrated with a Raspberry Pi 4 were positioned on the torso of a Pollen Robotics Reachy robot to capture audio signals from various touch types on the arms. The nylon PA-12 casing influenced the acoustic properties of the signals, which were spread across a wide frequency range (500 Hz to 16 kHz), contrasting with human mechanoreceptor expectations (2 to 500 Hz) [4]. Preprocessing included DC offset removal, trimming, high-pass filtering (cutoff at 1 kHz), and normalization. These pre-processed signals were transformed into spectrograms, with extracted features such as duration, amplitude, RMS, frequency metrics, and energy distribution across multiple bands, forming the dataset, containing 336 samples in total (48 samples per touch type), which was used to train a CNN model [5] for touch classification.

## RESULTS AND DISCUSSION

Analysis of the audio signals revealed significant differences between raw and processed signals. The original signals showed strong DC offset and noise, while the processed signals demonstrated reduced noise and clearer touch frequencies (Fig 1), rendering them suitable for touch classification. Different types of touch were detected with the accuracies shown in Tab. 1.

Frequency analysis revealed that similar touch types, such as a "knock" and a "tap", or a "rub" and a "stroke" share similar dominant frequencies (1938 Hz and 1769 Hz, or 1663 Hz and 1605 Hz, respectively), making them difficult to differentiate. Despite this, the model achieved high accuracy for "Scratch" (1.00) and "Stroke" (0.85). However, the similarity between "Knock" and "Tap" was confirmed by the model's difficulty in distinguishing them, with accuracies of 0.69 and 0.67, respectively. "Press" was also harder to detect, with an accuracy of 0.65, due to minimal acoustic changes when pressure was sustained. Higher classification accuracy could be obtained by integrating similar touch types into one class (Tab. 1).

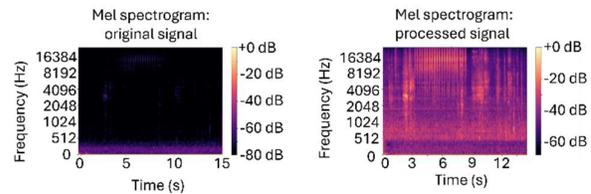

**Fig 1** Comparison of original and processed audio signals, showing reduced noise and clearer frequency content after preprocessing.

**Table 1:** Dominant frequencies and classification accuracy for different touch types

| Touch type | Dominant frequency (Hz) | Classification accuracy | Classification accuracy, integrated classes |
|---|---|---|---|
| Knock | 1938 ± 2143 | 0.69 | 0.81 |
| Tap | 1769 ± 2162 | 0.67 | |
| Rub | 1663 ± 2157 | 0.75 | 0.89 |
| Stroke | 1605 ± 1404 | 0.85 | |
| Scratch | 1641 ± 576 | 1 | 0.85 |
| Press | 2269 ± 3428 | 0.65 | 0.67 |

## CONCLUSIONS

The microphones successfully detected various touch types on the robot, but accurately classifying "Press" interactions was difficult due to minimal acoustic changes. A larger dataset is needed to improve classification. Future work will also focus on sound localization methods for an articulated robot and evaluating the force associated with different touches.